\documentclass[a4paper,fleqn]{cas-dc}
\usepackage{natbib}
\usepackage{subfig}
\usepackage{hyperref}       % hyperlinks
\usepackage{url}            % simple URL typesetting
\usepackage{amsfonts}       % blackboard math symbols
\usepackage{nicefrac}       % compact symbols for 1/2, etc.
\usepackage{microtype}      % microtypography
\usepackage{xcolor}         % colors
\usepackage{amsmath}
\usepackage{amssymb}
\usepackage{graphicx}
\usepackage{tabularx}
\usepackage{tabulary}
\usepackage{multirow}
\usepackage{paralist}
\usepackage{enumitem}
\usepackage{booktabs}
\usepackage{caption}
\usepackage{bm}
\usepackage[switch]{lineno}
\usepackage[ruled,linesnumbered,lined,boxed,commentsnumbered]{algorithm2e}

\newcommand{\tablestyle}[2]{\setlength{\tabcolsep}{#1}\renewcommand{\arraystretch}{#2}\centering\footnotesize}

\def\tsc#1{\csdef{#1}{\textsc{\lowercase{#1}}\xspace}}
\tsc{WGM}
\tsc{QE}
\tsc{EP}
\tsc{PMS}
\tsc{BEC}
\tsc{DE}
\begin{document} %avoid word overflow
%\linenumbers 
%\captionsetup{labelfont={rm}}
\begin{sloppypar}
\let\WriteBookmarks\relax
\def\floatpagepagefraction{1}
\def\textpagefraction{.001}
\shorttitle{ISPRS Journal of Photogrammetry and Remote Sensing}
\shortauthors{H. Zhu, et~al.}

\title [mode = title]{Robust Tiny Object Detection in Aerial Images amidst Label Noise}                      

\author[1]{Haoran Zhu}[style=chinese]
                        %auid=000,bioid=1,
                        %prefix=Sir,
                        %role=Researcher,
                        %orcid=0000-0001-7511-2910]
\cormark[1]
\ead{zhuhaoran@whu.edu.cn}
%\ead[url]{www.cvr.cc, cvr@sayahna.org}

%\credit{Conceptualization of this study, Methodology, Software}

\address[1]{School of Electronic Information, Wuhan University, Wuhan 430072, China}
\address[2]{School of Computer Science and the State Key Lab. LIESMARS, Wuhan University, Wuhan, 430072, China}

\author[1]{Chang Xu}[style=chinese,orcid=0000-0002-3078-0496]
\cormark[1]
\ead{xuchangeis@whu.edu.cn}

%\fnmark[2]

%\credit{Data curation, Writing - Original draft preparation}
\author[1]{Wen Yang}[style=chinese,orcid=0000-0002-3263-8768]
\cormark[2]
\ead{yangwen@whu.edu.cn}

\author[1]{Ruixiang Zhang}[style=chinese,orcid=0000-0003-0704-2484]                    
\ead{zhangruixiang@whu.edu.cn}

\author[1]{Yan Zhang}[style=chinese,orcid=0000-0003-4794-6082]                    
\ead{zhangyan@whu.edu.cn}

\author[2]{Gui-Song Xia}[style=chinese,orcid=0000-0001-7660-6090]
\ead{guisong.xia@whu.edu.cn}

\cortext[cor1]{Equal contributions}
\cortext[cor2]{Corresponding author}

\begin{abstract}
Precise detection of tiny objects in remote sensing imagery remains a significant challenge due to their limited visual information and frequent occurrence within scenes. This challenge is further exacerbated by the practical burden and inherent errors associated with manual annotation—annotating tiny objects is laborious and prone to errors (\textit{i.e.}, label noise). Training detectors for such objects using noisy labels often leads to suboptimal performance, with networks tending to overfit on noisy labels. In this study, we address the intricate issue of tiny object detection under noisy label supervision. We systematically investigate the impact of various types of noise on network training, revealing the vulnerability of object detectors to class shifts and inaccurate bounding boxes for tiny objects. To mitigate these challenges, we propose a DeNoising Tiny Object Detector (DN-TOD), which incorporates a Class-aware Label Correction (CLC) scheme to address class shifts and a Trend-guided Learning Strategy (TLS) to handle bounding box noise. CLC mitigates inaccurate class supervision by identifying and filtering out class-shifted positive samples, while TLS reduces noisy box-induced erroneous supervision through sample reweighting and bounding box regeneration. Additionally, Our method can be seamlessly integrated into both one-stage and two-stage object detection pipelines. Comprehensive experiments conducted on synthetic (\textit{i.e.}, noisy AI-TOD-v2.0 and DOTA-v2.0) and real-world  (\textit{i.e.}, AI-TOD) noisy datasets demonstrate the robustness of DN-TOD under various types of label noise. Notably, when applied to the strong baseline RFLA, DN-TOD exhibits a noteworthy performance improvement of 4.9 points under 40\% mixed noise. Datasets, codes, and models will be made publicly available.
\end{abstract}


\maketitle
% introduction
\section{Introduction}
Tiny objects, characterized by their extremely limited appearance information (less than 16$\times$16 pixels), are scattered across aerial images captured from different heights and locations. The accurate detection of tiny objects is always one of the primary challenges on the way towards automatic interpretation of aerial images, drawing increasing attention in both computer vision and remote sensing communities~\citep{rfla,aitodv2_2022_isprs,dotav2,soda_2023_pami,dcfl_cvpr_2023,TinyPerson_2020_WACV,R2CNN_2019_TGRS}.

In response, much effort has been devoted to the establishment of Tiny Object Detection (TOD) datasets~\citep{TinyPerson_2020_WACV,AI-TOD_2020_ICPR,aitodv2_2022_isprs,dotav2,soda_2023_pami,manipal_2023_isprs} and customized object detectors~\citep{rfla,aitodv2_2022_isprs,querydet_2022_cvpr}.
In the data-driven deep learning era, TOD presents unique difficulties not only for existing detectors but also for the construction of datasets. Tiny objects' lack of appearance makes the labeling process quite laborious and easy to introduce label noise.
While prior research, exemplified by AI-TOD-v2.0~\citep{aitodv2_2022_isprs}, has shed light on challenges such as missing labels, it has not directly addressed these issues. In this work, we delve deeper into label noise in tiny object detection and take a step forward to handle this issue by first analyzing the impact of various kinds of label noise and then proposing a robust detector capable of mitigating the impact of label noise.

\begin{figure}[t]
    \centering
    %\subfigure{
    \includegraphics[width=0.99\linewidth]{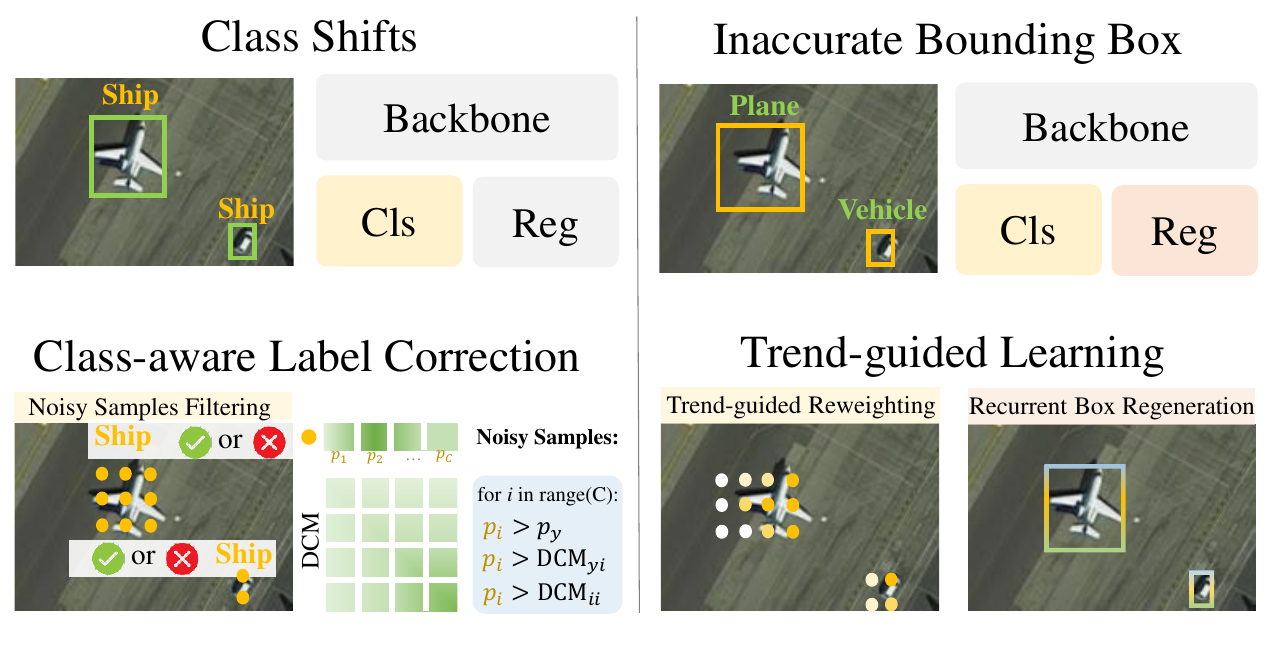}
    %}
    \caption{ An overview of our method. Observing that the model is quite sensitive to \textit{class shifts} and \textit{inaccurate bounding box} noises when training tiny object detectors, we propose to tackle them by Class-aware Label Correction and Trend-guided Learning, respectively. Orange points denote positive samples.}
    \label{fig:overall}
    %\vspace{\fixedvskip}
\end{figure}

Compared to image classification, label noise types in object detection are more diverse and complex. They can be classified into four types, namely missing labels, extra labels, class shifts, and inaccurate bounding boxes. Some previous studies~\citep{bernhard2021correcting, chadwick2019training, li2020towards, metalearning} assume all types of noise occur and try to tackle all types of noise simultaneously, while others~\citep{SDlocnet, OAMIL, SSDdet} focus on handling a specific type of noise (\textit{e.g.} inaccurate bounding box)~\citep{OAMIL,SSDdet}. 
Considering that tiny objects exhibit unique characteristics different from generic objects~\citep{AI-TOD_2020_ICPR}, we focus on addressing label noise in tiny object detection. First of all, we systemically investigate the detector's performance under different kinds of label noises (\textit{i.e.}, missing labels, extra labels, class shifts, inaccurate bounding boxes). Then, we find out that two types of noises will significantly reduce the detection performance, namely the \textbf{class shifts} and \textbf{inaccurate bounding boxes}.

To mitigate the impact of these two types of noises, we propose a DeNoising Tiny Object Detector (DN-TOD). However, distinct challenges are raised when tackling class shifts and inaccurate bounding boxes. 
The class imbalance issue in tiny object detection datasets can make frequent classes yield significantly higher confidence than rare classes. Consequently, class-agnostic methods~\citep{labelnoiseclassimbalance,coteaching} perform poorly in addressing mislabeled samples of the rare classes.

Besides, inaccurate bounding boxes can affect both the classification and regression. Similar to previous studies on the label noise of generic object detection~\citep{OAMIL}, we also observe that tiny object regression is quite vulnerable to bounding box offsets. Tiny objects extend challenges to the classification branch, bounding box offsets from objects' main body will deteriorate tiny objects' lack of high-quality positive samples~\citep{rfla}, simultaneously reducing classification accuracy. 

DN-TOD addresses the class shifts and inaccurate bounding boxes with the Class-aware Label Correction (CLC) scheme and the Trend-guided Learning Strategy (TLS), respectively. An overview of our approach is depicted in Fig.~\ref{fig:overall}. Specifically, the CLC module can be separated into the class confusion state updating and noisy sample filtering process. In the updating stage, we design a class-aware Dynamic Confidence Matrix (DCM). The DCM is updated by network prediction of different classes over a period. This matrix dynamically encodes the transfer probability between different classes, providing class discrimination criteria for the filtering stage under class imbalance conditions. In the filtering stage, we traverse positive sample's each class prediction and identify noisy samples through three heuristic rules built upon the DCM and predicted confidence. We discard the learning of identified noisy samples to filter out inaccurate supervision information.

\begin{figure}[t]
    \centering
    \includegraphics[width=0.99\linewidth]{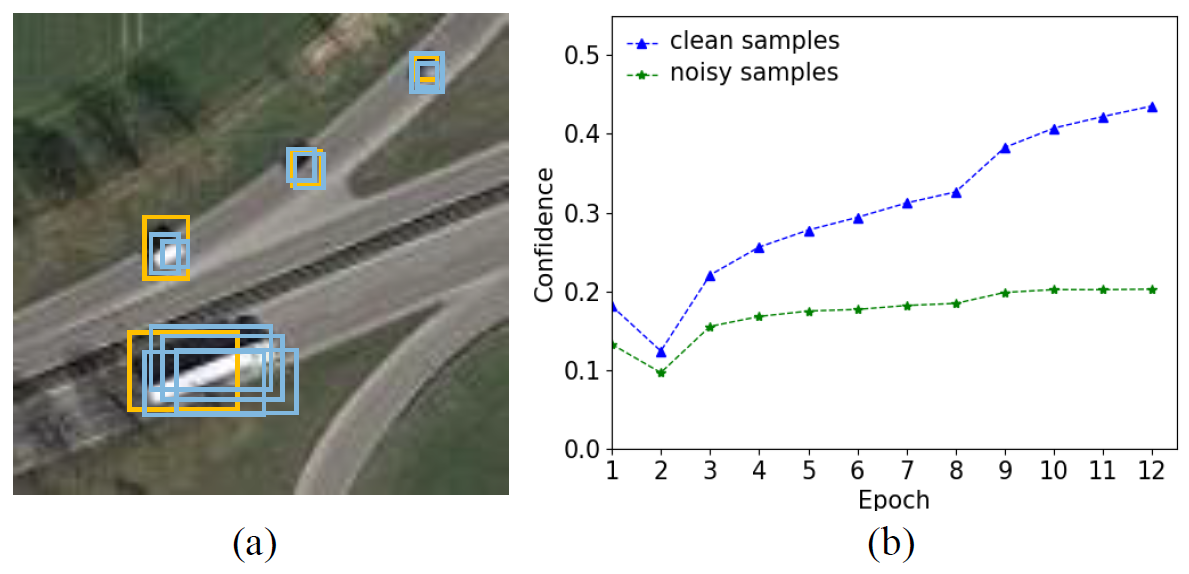}
    \caption{ (a) illustrates the noisy box labels \textit{vs.} predicted boxes. The orange box represents inaccurate bounding box annotations. The blue box represents the prediction boxes of the regression branch. The network can still provide relatively accurate regression predictions even under noisy box supervision. (b) stands for the confidence changing trend at different epochs. Clean samples exhibit an upward learning trend, while noisy samples show a downward and constant trend during training.}
    \label{fig:prediction trend}
\end{figure}

On the other hand, in light of the dual effects brought by inaccurate bounding boxes, we propose to bolster the detector's robustness of classification and regression simultaneously. Our design is based on a key observation of the learning trend difference between clean and noisy samples. As shown in Fig.~\ref{fig:prediction trend}(b), we count the mean confidence score of clean and noisy samples in the synthesized noisy dataset 
and find out that clean samples exhibit a gradually increasing trend, while the confidence of noisy samples will fluctuate within a low score range during training. 

Thus, the learning trend serves as a convincing element to evaluate the noisy condition of samples in the dataset. Instead of setting a hard threshold and separating samples from non-noisy or clean, we use the trend as soft labels to refactor the supervision target when training.
For classification, we design a Trend-guided Reweighting Loss, which enhances clean positive samples while reducing the gradient contribution of noisy samples. 
For regression, we rectify the regression target via a Recurrent Box Regeneration strategy. Indeed, the deep neural network has a certain robustness to label noise, as shown in Fig.~\ref{fig:prediction trend}(a), the network can make predictions more accurate than the noisy ground truth (\textit{gt}) during training. To avoid the network's overfit to noisy boxes, we rectify the box label via selectively ensemble predictions of clean samples and the original \textit{gt}.

The contributions of this paper are three-fold: 

\begin{itemize}
    \item We investigate the impact of different types of label noise in tiny object detection, and highlight two types of noise that severely impair the detection performance: class shifts and inaccurate bounding boxes.
    \item We propose a DeNoising Tiny Object Detector (DN-TOD) that performs robust object detection under label noise. DN-TOD is composed of a Class-aware Label Correction (CLC) module for mitigating class shifts and a Trend-guided Learning Strategy (TLS) for rectifying box noise.
    \item Experiments on synthesized label noise aerial object detection datasets (\textit{i.e.}, AI-TOD-v2.0 and DOTA-v2.0) and real-world label noise (\textit{i.e.}, AI-TOD) demonstrate our DN-TOD outperforms previous methods, achieving state-of-the-art (SOTA) results.
\end{itemize}

The rest of this paper is organized as follows. In Section~\ref{relatedwork}, we make a brief survey of related works. In Section~\ref{Preliminary}, we introduce the definition and influence of label noise on tiny object detection and the construction method of the synthetic noisy dataset. Section~\ref{methdo_overall} shows details of DN-TOD. Then, we validate the effectiveness of the proposed method under different types of noises in Section~\ref{experiments} and provide a detailed discussion in Section~\ref{discussion}. Finally, we conclude this paper in Section~\ref{conclusion}.

% related work
\section{Related Work}
\label{relatedwork}
\subsection{Tiny Object Detection}
The extremely limited number of pixels severely degrades the generic object detectors' performance on tiny object detection, stimulating a growing trend of specialized studies. In short, we can distinguish methods designed for detecting tiny objects as follows.

Multi-scale image or feature representations are classic strategies to tackle small or tiny objects. At the image level, SNIP~\citep{SNIP_2018_CVPR} and SNIPer~\citep{SNIPER_2018_NIPS} normalize the objects' scale in a specific range for scale-invariant detection. FPN serves as the cornerstone in the feature-level multi-scale strategy~\citep{FPN_2017_CVPR,hynet_2021_isprs,msod_2018_isprs}, followed by its variants like PANet~\citep{PANet_2018_CVPR}, Recursive-FPN~\citep{Efficientdet_2020_CVPR}, BiFPN~\citep{DetectoRS_2020_CVPR}, and the head-based TridentNet~\citep{Trident-Net_2019_ICCV}. 
Similar to multi-scale operations, super-resolution-based detectors are powerful for enhancing the feature representation of tiny objects~\citep{SOD-MTGAN_2018_ECCV,PGAN_2017_CVPR,Better_to_Follow_2019_ICCV,msod_2018_isprs}. 
Recent works cast new insights into the label assignment of TOD. Observing the IoU's vulnerability to box offsets, some works design new metrics for higher quality assignment~\citep{aitodv2_2022_isprs,dotd_2021_cvprw,bdistance_grsl_2022}. More essentially, recent works~\citep{rfla,dcfl_cvpr_2023} (\textit{e.g.}, RFLA) design scale-balanced assignment strategies to obtain sufficient supervision for tiny objects. 

Previous works have significantly pushed forward the TOD under the hypothesis of a clean training set. However, in the real world, TOD faces severe label noise problems. This work, instead, pursues robust learning for TOD under label noise.

\subsection{Learning with Noisy Labels}
Learning with noisy labels is always a research hotspot. Early works primarily focused on image classification tasks and proposed a series of noise-robust training methods~\citep{confidencelearning, coteaching, JoCoR, SimCLR, ma2018dimensionality} to address label noise issues. Specifically, confidence learning~\citep{confidencelearning} cleans noisy samples by estimating the joint distribution of noisy and clean labels. Sample selection-based methods such as Co-teaching~\citep{coteaching} and JoCoR~\citep{JoCoR} consider samples with smaller losses as clean and utilize them for network training. SimCLR~\citep{SimCLR} and MoCo~\citep{MoCo} employ contrastive regularization functions and architectures to learn representations of clean and noisy samples, aiding in the identification of noisy labels.

Recently, learning with noisy labels in object detection has gained a growing interest in the community. Much effort has been devoted to alleviating the effect of noisy labels. GFL~\citep{GFL} models bounding boxes as arbitrary distributions instead of Dirac distributions, efficiently guiding localization quality estimation through the uncertainty of bounding boxes. Wise-IoU loss~\citep{wiseiou} uses a dynamic non-monotonic focus mechanism that assesses anchor box quality with outliers.
MRNet~\citep{metalearning} utilizes a meta-learning-based model that addresses label noise problems through alternating noise correction and model training. Motivated by the fact that object detectors are more vulnerable to inaccurate bounding box annotations, some studies concentrate on tackling box noise.
OA-MIL and SSD-Det~\citep{OAMIL, SSDdet} use Multiple Instance Learning (MIL) based methods for robust learning under box noise. They construct high-quality proposal bags and filter out the finest proposals from the bags to serve as new labels.

As types of label noise in object detection are diverse, and different types of label noise degrade object detection in divergent ways, addressing the label noise in object detection is more complex and annoying than that of image classification. In this study, before the designing of noise-robust methods, we first investigate the impact of different types of noise in tiny object detection. Then, we focus on two types of noise that significantly reduce the detection performance of tiny objects and propose a DN-TOD that pursues noise-robust learning.
 
% preliminary
\section{Preliminary}
\label{Preliminary}
For a better understanding of label noise types in object detection, we provide their definitions first. Then, we investigate the impact of different types of label noise on tiny object detection by pilot experiments on the AI-TOD-v2.0.

\begin{figure}[t]
    \centering
    \includegraphics[width=0.95\linewidth]{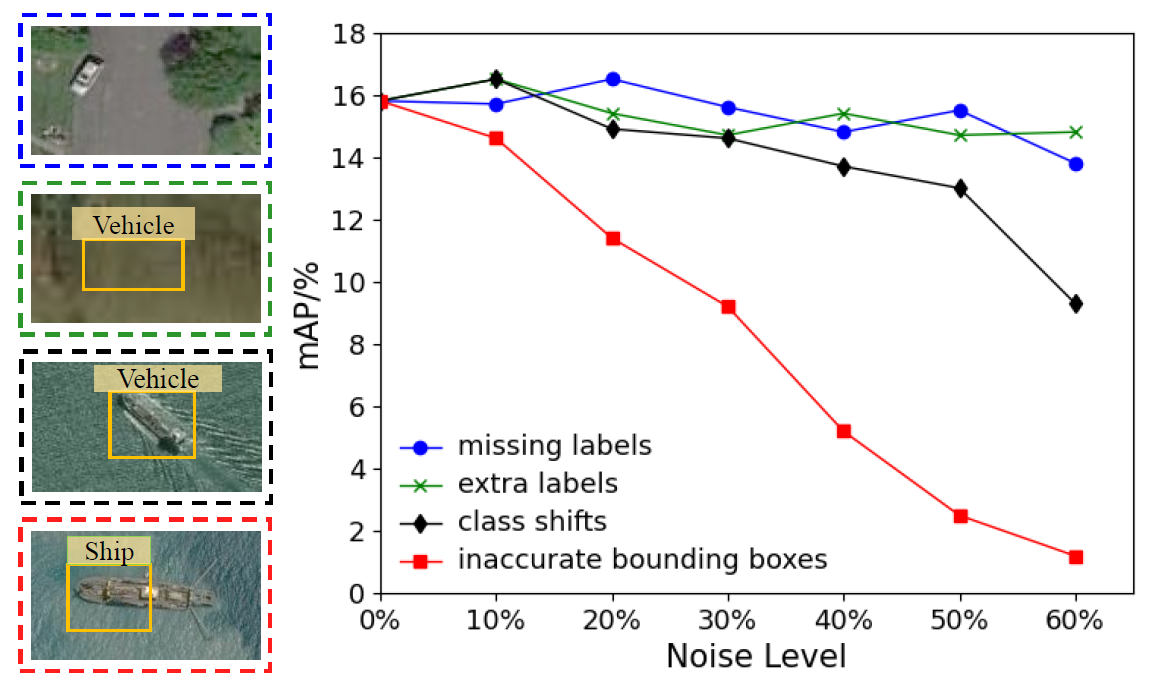}
    \caption{Mean average precision mAP@[.5,.95] of FCOS with RFLA on the synthesized “noisy” AI-TOD-v2.0 dataset, where the annotations are randomly perturbed. With the noise
level increases, \textit{i.e.}, annotations become more and more inaccurate, the mAP of class shifts and inaccurate bounding boxes drops significantly while the mAP of missing labels and extra labels still maintains high accuracy. }
    \label{fig:noise_influence}
    %\vspace{\fixedvskip}
\end{figure}

\subsection{Label Noise Categorization in Object Detection}
Label noise is defined as the disparity between the label instances used for training and the real object instances. Label noise in object detection can be separated into four types shown in Fig.~\ref{fig:noise_influence}, \textit{i.e.}, missing labels, extra labels, class shifts, and inaccurate bounding boxes. 
We define regions of interest missed to be annotated as \textit{missing labels}, and define regions out of interest that are wrongly labeled with foreground classes as \textit{extra labels}. 
When a foreground object is annotated with a wrong class label, it is regarded as the \textit{class shift}. When the provided bounding box coordinates offset from the object's main body, this annotation is defined as an \textit{inaccurate bounding box}.

\begin{figure}[t]
    \centering
    \includegraphics[width=0.95\linewidth]{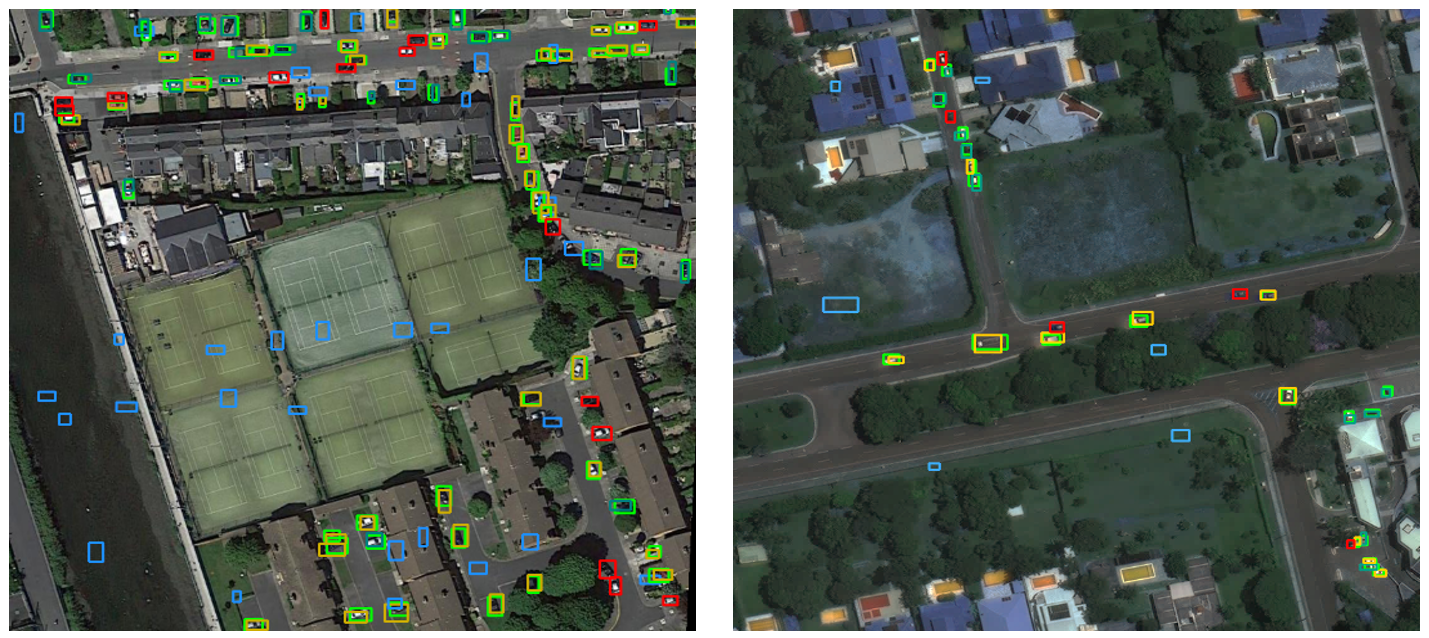}
    \caption{ 30\% simulated “noisy” AI-TOD-v2.0 dataset (green for original clean labels, red for missing labels, blue for extra labels, yellow for inaccurate bounding boxes, and dark cyan for class shifts). }
    \label{fig:simulated sample}
    %\vspace{\fixedvskip}
\end{figure}

\subsection{Noisy Tiny Object Detection Dataset Synthesis}
\label{dataset_syn}
Based on the definition of four types of label noise, we generate a series of simulated “noisy” datasets to ablate the effect of each type of noise. A dataset of noise level “$a\%$” is synthesized via randomly sampling “$a\%$” objects and posing permutations to them.
The core operation when synthesizing class noise is to switch classes, including foreground and background classes.
Depending on the specific type of class noise, it can be further classified as follows. For missing labels, we set the class of sampled objects to the background. Likewise, we randomly switch the class labels of sampled objects to other foreground classes to synthesize class shifts. 
Besides, for extra labels, we generate random bounding boxes at random positions on the image, annotating with random foreground labels. 

On the other hand, we follow the OA-MIL~\citep{OAMIL} to synthesize box noise. We simulate inaccurate bounding boxes by randomly perturbing the annotations of each clean box in the dataset, which can be written as:
\begin{equation}
    \begin{cases}
    \hat{c x}=c x+\Delta_x \cdot w, & \hat{c y}=c y+\Delta_y \cdot h, \\ \hat{w}=\left(1+\Delta_w\right) \cdot w, & \hat{h}=\left(1+\Delta_h\right) \cdot h,
    \end{cases}
\end{equation}
where $\Delta_x, \Delta_y, \Delta_w$, and $\Delta_h$ are randomly sampled from the uniform distribution $U(-a\%, a\%)$ when synthesizing a dataset of box noise level $a\%$.

The regression noise assumes the absence of clean bounding boxes while classification noise contains a certain proportion of clean labels. We visualize some examples of the simulated noisy AI-TOD-v2.0 in Fig.~\ref{fig:simulated sample}.

\subsection{The Impact of Label Noise}
\label{sec:impact}
To thoroughly investigate the impact of label noise on the tiny object detection task, we test the detection performance of different noise levels under each type of noise on the noisy AI-TOD-v2.0 dataset.
Experimental results are shown in Fig.~\ref{fig:noise_influence}. 
For all kinds of noise, the network's performance initially shows a trend of fluctuation as we start to add noise into the dataset.  
As the noise level further increases, the detection performance develops in divergent ways for different types of noise. The detector is relatively robust to issues related to missing labels and extra labels for tiny objects, as the mAP is reduced by less than 2 points as the noise increases to 60\%. However, the detector is quite vulnerable to class shifts and inaccurate bounding boxes, especially, the mAP is reduced by nearly half when there is 30\% box noise in the dataset. 

Based on these observations, we propose a robust detector to address issues that have a greater impacts on the tiny object detector in the following section (\textit{i.e.}, the class shifts and inaccurate bounding boxes).

% method
\section{Methodology}
\label{methdo_overall}
As mentioned in Section~\ref{sec:impact}, problems such as class shifts and inaccurate bounding boxes can result in a sharp decline in network performance. Thus, in this section, we introduce the DN-TOD designed for these two types of noise. Specifically, to address the class shifts, we design a Class-aware Label Correction scheme to discard the wrong category. Also, we propose a Trend-guided Learning Strategy to mitigate the impact of inaccurate bounding boxes on classification and regression branches. 

\subsection{Class-aware Label Correction}
\label{CLC}
Typical tiny object detection datasets include both frequently appearing classes (\textit{e.g.}, vehicle and ship) and rare classes (\textit{e.g.}, wind-mill and helicopter). When the TOD's weak appearance information is coupled with the extremely imbalanced class distribution, detectors will exhibit severe confidence bias towards the frequently appearing classes. 

One simple yet effective way of class label correction in image classification is to rectify the one-hot targets via the most confident class predicted by the model~\citep{regularizedestimation_cvpr_2019}. However, this strategy will exacerbate the class shift issue by misleading massive class targets to the head class (\textit{e.g.}, vehicle), since they tend to yield dominant confidence among all classes.

\begin{figure}[t]
    \centering
    \includegraphics[width=0.99\linewidth]{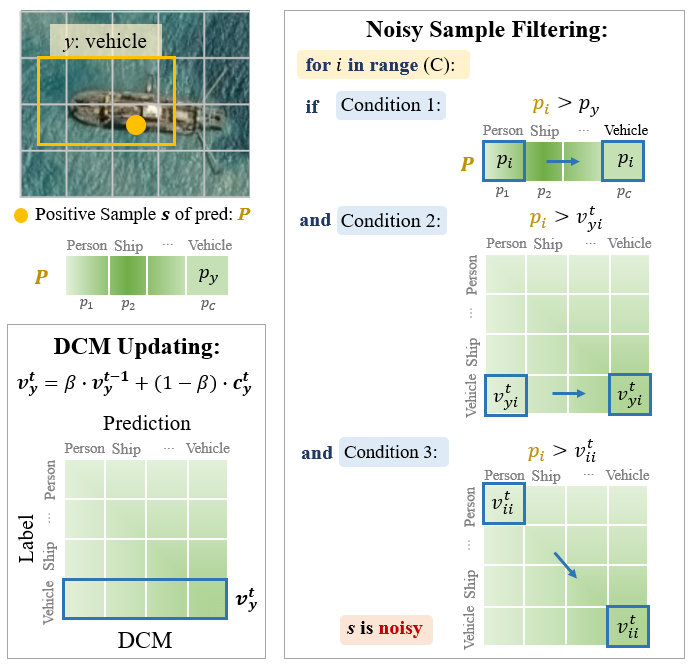}
    \caption{The workflow of Class-aware Label Correction (CLC). The class-aware label correction can be separated into the class confusion state updating and noisy sample filtering process. The DCM is updated by Eq.~\ref{con:update}, which is updated when a new image comes. 
    For each positive sample, we compare the predicted values of all classes $P$ with the DCM and determine whether this sample is noisy based on Eq.\ref{con:Gamma}. If any class prediction $p_i$ simultaneously satisfies the three conditions, the positive sample is considered noisy.}
    \label{fig:CLC}
    %\vspace{\fixedvskip}
\end{figure}

To this end, we propose a class-aware label correction scheme that sets dynamic thresholds for different classes. Followed by a simple yet effective scheme for filtering out noisy samples, the class-aware manner avoids the detector's over-confidence in classes of large quantity. Concretely, the class-aware label correction can be separated into the class confusion state updating and noisy sample filtering process. The workflow of CLC is shown in Fig.~\ref{fig:CLC}.

\textbf{Updating}. We design a Dynamic Confidence Matrix (DCM), which preserves and updates the class confidence state over a specific training time. 
The DCM ($D\in \mathbb{R}^{2}$) is a continuously updated tensor, where the first dimension is the label space, and the second dimension is the prediction space. $D$ is of size $\rm{C\times C}$, $\rm{C}$ is the class number.
As long as the initialization of $D$, it will be updated whenever a new image is sent to the pipeline. 
Specifically, given an image, we calculate the confidence pillar $c_i$ ($\rm{1\times C}$) of each \textit{gt} class, which is the average prediction of all positive samples belonging to class $i$ in this image. 
We then leverage \textit{Exponential Weighted Moving Average} to update the $y^{th}$ ($y\in \{1, 2, ..., C\}$) row of DCM with the $c_y$ pillar. 
The specific update pipeline can be represented as follows:
\begin{equation}
    v_{y}^{t} = \beta \cdot v_{y}^{t-1} + (1 - \beta) \cdot c_y^t, 
    \label{con:update}
\end{equation}
\begin{equation}
    \beta = 1-\frac{1}{T},
    \label{con:betatoT}
\end{equation}
where $v_{y}^{t-1}$ represents the vector value of the $y^{th}$ row in DCM at $t-1$ time, $c_y^t$ represents the average prediction of all positive samples belonging to class $y$ at $t$ time, and $v_{y}^{t}$ represents the updated vector value. It is worth noting that they are all $\rm{1\times C}$ vectors.
$\beta$ is a hyperparameter that denotes the weighting coefficient. To better interpret $\beta$, we use a more understandable $T$ which indicates taking the average value of the previous $T$ moments.

\textbf{Filtering}. After a period of warm-up, we ensemble the information across previous training time. 
Given the class-aware prediction $P =\{p_1,..p_C\}$ of a positive sample $s$ with its \textit{gt} label $y$, we propose a simple method to identify whether it is noisy or not. The sample-wise noisy state is identified by a noisy factor $D$, which can be expressed as follows:
\begin{equation}
    D(s) = \prod_{i=1}^C \big(1 - \epsilon(p_i - p_y) \cdot \epsilon(p_i - {v}_{yi}^t) \cdot \epsilon(p_i - {v}_{ii}^t)\big),
    \label{con:Gamma}
\end{equation}
where $v_{ij}^t$ denotes the element in the $i^{th}$ row and $j^{th}$ column of DCM. $\epsilon$ denotes the step function (\textit{i.e.}, if the input is greater than 0, the output is 1; otherwise, the output is 0). Finally, we disable the learning of the identified noisy samples via incorporating $D$ into the classification loss, which is written as:
\begin{equation}
    L_{cls} = \sum_{m=1}^{N_{pos}} D(s_m) \cdot FL(P_m, y_m) + \sum_{n=1}^{N_{neg}} FL(N_n, 0),
    \label{con:cls}
\end{equation}
where $N_{pos}, N_{pos}$ denote the number of positive samples and negative samples respectively. $FL(\cdot)$ denotes the Focal loss. $P_m, N_n$ are class-aware predictions of positive samples and negative samples respectively. $y_m$ denotes the annotated \textit{gt} label.

\begin{figure*}[t]
    \centering
    \includegraphics[width=0.99\linewidth]{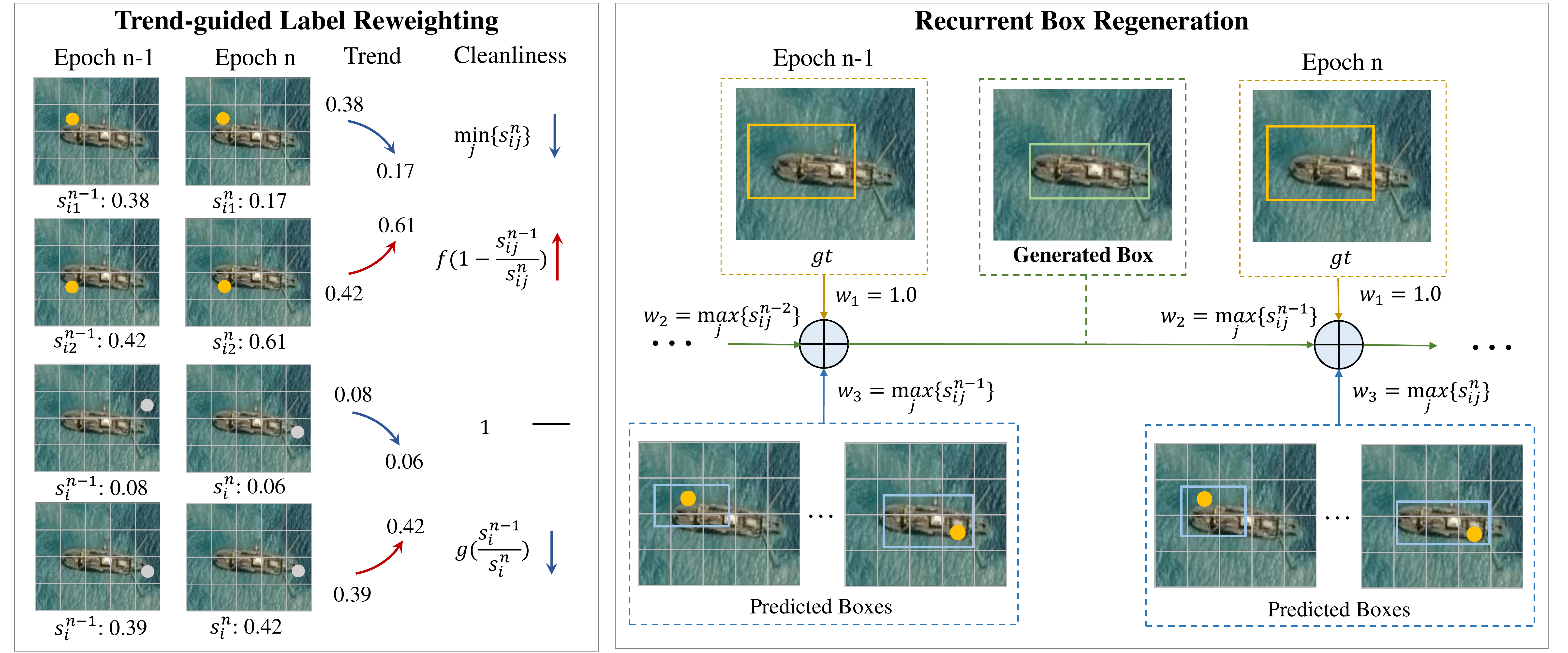}
    \caption{The overall process of Trend-guided Learning Strategy (TLS). TLS comprises two sub-modules: Trend-guided Label Reweighting (TLR) and Recurrent Box Regeneration (RBR).
    TLR is designed for the classification branch and assigns a cleanliness score to each sample based on its learning trend. RBR is designed for the regression branch by fusing the prediction results of all previous rounds to obtain a cleaner \textit{gt} box. Orange and gray points denote positive and negative samples respectively.}
    \label{fig:TLS}
    %\vspace{\fixedvskip}
\end{figure*}

\subsection{Trend-guided Learning Strategy}
As previously mentioned, for tiny object detection, inaccurate bounding boxes can simultaneously affect both the classification and regression branches. Therefore, we simultaneously handle the effects of inaccurate bounding boxes on the classification and regression via a Trend-guided Learning Strategy (TLS). TLS primarily comprises two sub-modules: Trend-guided Label Re-weighting (TLR) and Recurrent Box Regeneration (RBR), which rectify the supervision signals on the classification and regression branches respectively. 

\subsubsection{Trend-guided Label Reweighting}
\label{method:tlr}
Inaccurate bounding boxes can further exacerbate the lack of high-quality positive samples for tiny objects, providing misleading supervision signals for the classification branch. To rectify the positive and negative training samples for tiny objects, we introduce a Trend-guided Label Reweighting approach. 
Further, we assign weights to each sample based on its learning trend, termed as cleanliness. For positive samples, we enhance the learning of positive samples with a consistent growth trend and reduce the effects of uncertain samples during training. The opposite holds for negative samples.

Specifically, we note the positive sample set assigned to the $gt_i$ as $\rm{Pos}_i$. For each sample with a predicted score $s_{ij}$ in this set, we simply determine its cleanliness weight by the learning trend in different epochs as follows:
\begin{equation}
    %T\!rend_i^{pos}=f(1-\frac{p_j^{t-1}}{p_j^t})
    c_{ij} = f\big(1-\frac{s_{ij}^{n-1}}{s_{ij}^n}\big),
\end{equation}
where $n$ denotes the training epoch. Note that $f(x)=x, x\geq 0$, in other words, function $f$ is applicable only if the input value is not lower than zero. Otherwise, we set the $c_{ij}$ to the lowest positive value held by samples in $\rm{Pos}_i$.

During the initial stages of training, the network's predictions are often unstable, and the $c_{ij}$ tends to exhibit randomness. To ensure that each \textit{gt} can be stably assigned with at least one high-weight positive sample, we also introduce the concept of sample primacy when performing reweighting, which is written as:

\begin{equation}
    r_{ij}=\frac{s_{ij}}{\sum_{j=1}^{N_i} s_{ij}},
\end{equation}
where $N_i$ is the number of positive samples assigned to $gt_i$. For short, the loss weight for each positive sample is defined as:
\begin{equation}
    w_{ij}^{pos} = \alpha \cdot c_{ij}+(1-\alpha) \cdot r_{ij},
\end{equation}
in which $\alpha$ is a tunable hyperparameter.

Likewise, we also rectify the supervision signal provided by negative samples via their learning trend. We down-weight negative samples showing a growing trend since they are more likely to fall on objects' real bodies, which are wrongly assigned as negative samples owing to the inaccurate bounding boxes.  

Therefore, for each negative sample, we determine its loss weight by:
\begin{equation}
    w_i^{neg}=g\big(\frac{s_i^{n-1}}{s_i^n}\big), \quad
     g(x) = \begin{cases}
    1, & x>1 \\
    x, & x \leq 1 
    \end{cases} ,
\end{equation}

\subsubsection{Recurrent Box Regeneration}
\label{method:rbr}
Previous works have demonstrated that the temporal self-ensembling of network parameters or predictions can make the model more robust to label noise~\citep{selfensemble}.
In this paper, we extend this idea to tiny object detection and design a Recurrent Box Regeneration (RBR) strategy. The core idea of RBR is to recurrently leverage previous rounds' predictions and \textit{gt} to rectify the noisy box learning target when training.
In this approach, the previous round's bounding box predictions are fused and weighted with the current round's prediction information to generate new supervision targets. However, purely relying on multi-round predictions as regression targets can result in unstable training. Therefore, we also incorporate \textit{gt} to obtain more stable and appropriate supervision for training.

Assuming that we obtain the top $k$ predictions for each \textit{$gt_i$} from the network, denoted as $B_i^n=\{(\bm{b}_{i1}^n,s_{i1}^n), \dots, (\bm{b}_{ik}^n,s_{ik}^n)\}$, where $B^n$ represents the $n^{th}$ epoch of network predictions, $\bm{b}$ represents the predicted bounding box coordinates (\textit{i.e.}, $\{ l,t,r,b \}$ in FCOS~\citep{fcos}), and $s$ represents the class confidence. We regenerate a new box target by fusing the predicted box from the current epoch, the box target from the previous epoch, and the \textit{gt} box. The specific fusion method is shown as follows:
\begin{equation}
    \bm{\theta}_i^n= w_1 \cdot \bm{g}_{i}+ w_2 \cdot \bm{\theta}_i^{n-1} + w_3 \cdot\sum_{j=1}^k s_{ij}^n \cdot \bm{b}_{ij}^n ,
    \label{con:rbrfusion}
\end{equation}
where $\bm{\theta}_i^n$ and $\bm{\theta}_i^{n-1}$ represent the $i^{th}$ box targets generated by the RBR module in the $n^{th}$ and ${(n-1)}^{th}$ epoch, respectively. $\bm{g}_i$ denotes the $i^{th}$ noisy \textit{gt} box, the $w_1$ is an adjustable super-parameter. The contributions of the RBR-generated box from the previous round and the predicted boxes are obtained by:
\begin{equation}
    w_2 = \mathop{max}\limits_{j}\{ s_{ij}^{n-1} \}, w_3 = \mathop{max}\limits_{j}\{ s_{ij}^{n} \}.
    \label{con:w2}
\end{equation}

\subsection{Application to Object Detectors}
Our approach is a plug-and-play and can be deployed into various object detectors. 
We demonstrate the DN-TOD's versatility by plugging it into two representative baselines: the one-stage method FCOS, and the two-stage method Faster R-CNN. We use the RFLA strategy in both methods to provide strong baselines for tiny object detection. For simplicity, we note FCOS with RFLA and Faster R-CNN with RFLA as FCOS* and Faster R-CNN* respectively.

For the one-stage method FCOS*, all our designs (CLC, TLR, RBR) can be simply deployed onto the final detection head. In the case of Faster R-CNN*, our proposed method can be incorporated into both the first and second stages. In the first stage, since the RPN provides coarse proposals for the R-CNN stage, we utilize the TLR module that assesses the cleanliness of negative samples in addition to the RBR and CLC modules. In the second stage, we deploy all modules as usual.

% experiments
\renewcommand{\arraystretch}{1.2}
\begin{table*}[t]
	\centering
	\caption{Performance of baseline methods and our proposed method on the AI-TOD-v2.0 {\tt val set} under different levels of class shifts. * means using RFLA~\citep{rfla} in the label assignment. Abbreviations are used to define different categories.}
	\begin{tabular}{c | l | c c c c c c c c | c}%四个c代表该表一共四列，内容全部居中
        \toprule
        %\textbf{Noise Level} & \textbf{Method} & \textbf{VE} & \textbf{SH} & \textbf{PE} & \textbf{ST} & \textbf{SP}& \textbf{AI}& \textbf{BR}& \textbf{WH} & \textbf{mAP}\\
        Class Noise Level & Method & VE & SH & PE & ST & SP & AI & BR & WM & mAP \\
        \midrule
        \multirow{4}{*}{0\%} & FCOS*\citep{fcos} & 24.5 & 43.5 & 4.5 & 29.9 & 4.7 & 0.9 & \textbf{17.4} & 1.1 & 15.8\\
        & Faster R-CNN*~\citep{faster-rcnn} & \textbf{24.9} & 25.0 & 6.3 & 37.3 & \textbf{17.1} & \textbf{8.9} & 12.2 & 4.3 & 17.0\\

         & FCOS* w/ CLC  & 24.7 & \textbf{44.3} & 5.0 & 31.2 & 7.4 & 2.1 & 16.3 & 2.6 & 16.7\\
         & Faster R-CNN* w/ CLC  & 24.8 & 25.5 & \textbf{6.9} & \textbf{37.8} & 16.3 & 7.7 & 12.9 & \textbf{4.9} & \textbf{17.1} \\
        \midrule
        \multirow{4}{*}{10\%} & FCOS*  & 24.5 & 43.1 & 4.7 & 29.7 & 6.5 & 0.0 & 18.7 & \textbf{5.0} & 16.5\\
        & Faster R-CNN*  & 22.3 & 23.6 & 5.4 & 34.6 & 18.1 & 7.8 & 11.3 & 3.9  & 15.9\\

         & FCOS* w/ CLC  & \textbf{25.1} & \textbf{44.0} & 5.1 & 30.4 & 7.4 & 0.0 & \textbf{19.5} & 3.4 & 16.8\\
         & Faster R-CNN* w/ CLC  & 23.6 & 25.0 & \textbf{6.1} & \textbf{36.6} & \textbf{18.7} & \textbf{12.1} & 11.4 & 3.2 & \textbf{17.1} \\
        \midrule
        \multirow{4}{*}{20\%} & FCOS*  & 24.3 & 41.6 & 4.4 & 30.7 & 0.3 & 0.0 & 17.0 & 1.1 & 14.9\\
        & Faster R-CNN*  & 21.7 & 22.6 & 4.6 & 33.0 & 18.0 & 8.6 & 13.0 & \textbf{3.9}  & 15.7\\

         & FCOS* w/ CLC  & \textbf{24.5} & \textbf{43.2} & 4.6 & 30.7 & 0.2 & 0.0 & \textbf{18.7} & 2.7 & 15.6\\
         & Faster R-CNN* w/ CLC  & 23.5 & 24.4 & \textbf{5.9} & \textbf{36.7} & \textbf{20.5} & \textbf{9.6} & 11.0 & 3.1 & \textbf{16.9} \\
        \midrule
        \multirow{4}{*}{30\%} & FCOS* & 24.7 & 40.0 & 4.2 & 28.6 & 0.3 & 0.0 & 16.6 & 2.2 & 14.6 \\
        & Faster R-CNN*  & 21.3 & 23.5 & 5.2 & 31.6 & \textbf{20.0} & 7.6 & 9.4 &  3.7 & 15.3\\
         & FCOS* w/ CLC  & \textbf{24.7} & \textbf{43.3} & 4.7 & 29.9 & 0.1 & 0.0 & \textbf{17.0} & 0.0 & 15.0\\
         & Faster R-CNN* w/ CLC  & 23.7 & 24.8 & \textbf{5.9} & \textbf{37.1} & 19.6 & \textbf{8.8} & 11.6 & \textbf{3.9}  & \textbf{16.9}\\
        \midrule
        \multirow{4}{*}{40\%} & FCOS*  & 24.2 & 38.3 & 3.9 & 27.0 & 0.0 & 0.0 & 16.2 & 0.0 & 13.7\\
        & Faster R-CNN*  & 20.9 & 21.5 & 4.1 & 31.0 & 15.2 & 8.2 & 9.3 & \textbf{2.8} & 14.1\\
         & FCOS* w/ CLC  & \textbf{24.6} & \textbf{41.7} & 4.4 & 28.7 & 0.2 & 0.0 & \textbf{17.6} & 0.0  & 14.7\\
         & Faster R-CNN* w/ CLC  & 23.4 & 24.1 & \textbf{5.9} & \textbf{36.4} & \textbf{20.0} & \textbf{8.7} & 12.4 & 2.3 & \textbf{16.6} \\
        \bottomrule
        \end{tabular}
	\label{tab:classshifts}
\end{table*}

\section{Experiments}
\label{experiments}
\subsection{Experimental Settings}
\textbf{Synthetic Noisy Dataset.} 
To verify our method on tiny objects, we conduct main experiments and ablations on the challenging AI-TOD-v2.0~\citep{aitodv2_2022_isprs} dataset, which has the smallest average absolute object size of 12.7 pixels and contains 28, 036 images. We also verify the effectiveness of the proposed method on the generic aerial object detection scenario, with DOTA-v2.0~\citep{dotav2}.
We simulate noisy datasets following Sec.~\ref{dataset_syn}. Concretely, we simulate various noise levels ranging from 10\% to 40\% for the AI-TOD-v2.0 and DOTA-v2.0, respectively.

\renewcommand{\arraystretch}{1.3}
\begin{table*}
	\centering
	\caption{Comparison between other methods and our proposed TLS on the AI-TOD-v2.0 {\tt val set} with different levels of box noise. -- represents a failure to converge. * means using RFLA~\citep{rfla} for label assignment. We follow previous methods~\citep{OAMIL,SSDdet} and report $\rm{AP_{0.5}}$ in this table. }
	\begin{tabular}{l | l | c | c c c c}%四个c代表该表一共四列，内容全部居中
    \toprule
    \multirow{2}{*}{Stages} & \multirow{2}{*}{Method}& Clean & \multicolumn{4}{c}{Box Noise Level} \\%表格宽度参数采用*代表自动宽度
     & & 0\%  & 10\% & 20\% & 30\% & 40\% \\
    \midrule%第二道横线 
    \multirow{6}{*}{One-stage Methods} & FCOS* & 34.9 & 34.7 & 30.2 & 24.3 & 18.4 \\
    \cline{2-7}
    & Free Anchor~\citep{freeanchor} & 20.1 & 19.2 & 14.2 & 9.3 & 7.2 \\
    \cline{2-7}
    & Gaussian yolov3~\citep{gaussianyolov3} & 34.6 & 32.2 & 28.2 & -- & -- \\
    \cline{2-7}
    & Wise-IoU~\citep{wiseiou} & 35.7 & 36.6 & 30.8 & 23.7 & 16.8 \\
    \cline{2-7}
    & Generalized Focal Loss~\citep{GFL} & 37.6 & 36.9 & 33.9 & 26.6 & 14.9 \\
    \cline{2-7}
      & FCOS* w/ TLS & \textbf{41.0} & \textbf{39.8} & \textbf{37.4} & \textbf{32.1} & \textbf{21.3} \\
    \midrule
    \multirow{4}{*}{Two-stage Methods} & Faster R-CNN* & 45.0 & 44.0 & 39.3 & 29.0 & 15.4 \\
    \cline{2-7}
    & KL-Loss~\citep{klloss} & \textbf{48.8} & 45.7 & 38.2 & 26.5 & 15.9 \\
    \cline{2-7}
    & OA-MIL~\citep{OAMIL} & 46.0 & 42.7 & 39.3 & 28.5 & 15.0 \\
    \cline{2-7}
      & Faster R-CNN* w/ TLS & 47.3 & \textbf{46.1} & \textbf{39.5} & \textbf{31.9} & \textbf{19.2} \\
    %\midrule
    %\midrule
    %Clean Model & Clean-FCOS* & 34.9 & 34.9 & 34.9 & 34.9 & 34.9 \\
    %Clean Model & Clean-FasterRCNN* & 45.0 & 45.0 & 45.0 & 45.0 & 45.0 \\
    \bottomrule
    \end{tabular}
	\label{tab:inaccuratebbox}
\end{table*}
\textbf{Implementation Details.} 
We build the code using MMdetection~\cite{mmdetection_2019_arXiv} based on the PyTorch~\cite{PyTorch_2019_NIPS} deep learning framework. Similar to the default setting of object detection, the ImageNet~\citep{ImageNet_2015_IJCV} pre-trained model is used as the backbone. All models are trained using the Stochastic Gradient Descent (SGD) optimizer for 12 epochs with 0.9 momentum, 0.0001 weight decay, and 1 batch size. The initial learning rate is set to 0.005 and decays at the $8^{th}$ and $11^{th}$ epochs. Besides, the number of RPN proposals is set to 3000. In the inference stage, the confidence score is set to 0.05 to filter out background bounding boxes, and the NMS IoU threshold is set to 0.5 with the top 3000 bounding boxes. All the other parameters are set the same as default in MMdetection. The evaluation metric follows the AI-TOD-v2.0 benchmark~\citep{aitodv2_2022_isprs}, including AP, AP$_{0.5}$, AP$_{0.75}$, AP$_{vt}$, AP$_{t}$, AP$_{s}$, and AP$_{m}$. We set T to 100, $\alpha$ to 0.5, and $\omega$ from $\{1.0\dots 5.0\}$ (depending on datasets and noise levels). Besides, the filtering stage of CLC starts in the middle of the network training.

\subsection{Main Results}
For a clearer understanding of the effectiveness of each proposed strategy on the noise they target, we first investigate the effects of CLC on the class shifts and TLS on the box noise. Then, we verify the overall improvement brought by DN-TOD on the mixed noise and real-world noise settings. 

\textbf{Class Shifts.} The CLC module is designed to address the class shifts only. We compare our approach with the vanilla FCOS* and Faster R-CNN* models. As shown in Table~\ref{tab:classshifts}, our approach achieves consistent improvement over the vanilla models. The observed phenomenon echoes the previous claim that the class shift noise has a significant impact on the rare classes (\textit{e.g.} swimming-pool) while also confusing the other classes (\textit{e.g.} storage-tank). Notably, our CLC can offer substantial mitigation of the impact of class shifts on both frequently appearing classes and rare classes. Particularly, under 40\% noise, the CLC brings Faster R-CNN* back to the performance under the clean dataset, which only minorly drops by 0.5 points. 
Since some classes in AI-TOD-v2.0 (\textit{e.g.}, wind-mill) have very small numbers, the baseline performance is nearly random. Consequently, it is not hard to imagine that our method struggles to yield an obvious improvement in these classes.  

\textbf{Inaccurate Bounding Boxes.} Compared to other types of noise, bounding box noise is unique in the object detection task and will prominently affect the detection performance. Therefore, more previous works are focusing on mitigating the effects of this specific type of label noise.
Table~\ref{tab:inaccuratebbox} provides a comprehensive comparison of the proposed TLS with others on the AI-TOD-v2.0 under different levels of box noise. By categorizing studies into single-stage and two-stage methods, we have the following observations. First, two-stage methods are more robust to box noise on tiny objects. Our proposed method can both improve the one-stage and two-stage baselines by a large margin, and the improvement is particularly noticeable on the one-stage baseline, which lifts the baseline by 7.2 points under 20\% noise. Second, some methods designed to tackle the noise for generic object detection do not generalize well on detecting tiny objects in aerial images. For example, OA-MIL yields promising improvements on the noisy COCO dataset while it decreases by 1.3 points under 10\% noise compared to the baseline. This could be attributed to the learning bias towards larger objects when using predicted confidence to screen clean samples. By contrast, the learning trend in TLS eliminates bias in confidence-based methods towards larger objects, faring well on the robust detection of tiny objects across various levels of box noises.  

% \begin{table*}
% 	\centering
% 	\caption{Performance of baseline methods and our DN-TOD on the proposed AI-TOD-v2.0 {\tt val set} with different levels of mixed noise. * means using RFLA~\citep{rfla} in the label assignment.}
% 	\begin{tabular}{l | l | c c c c c}%四个c代表该表一共四列，内容全部居中
%         \toprule
%         \multirow{2}{*}{Model} & \multirow{2}{*}{Method}& \multicolumn{5}{c}{Noise Level} \\%表格宽度参数采用*代表自动宽度
%          & & 0\% & 10\% & 20\% & 30\% & 40\% \\
%         \midrule%第二道横线 
%         \multirow{2}{*}{FCOS*~\citep{fcos}} & Vanilla & 34.9 & 33.0 & 28.5 & 24.9 & 12.7 \\
%         \cline{2-7}
%         & DN-TOD & \textbf{38.1} & \textbf{34.2} & \textbf{31.4} & \textbf{27.2} & \textbf{17.6} \\
%         \midrule
%         \multirow{2}{*}{Faster R-CNN*~\citep{Faster R-CNN}} & Vanilla & 45.0 & 43.7 & 33.0 & 24.2 & 12.8 \\
%         \cline{2-7}
%           & DN-TOD & \textbf{45.5} & \textbf{45.3} & \textbf{38.1} & \textbf{29.1} & \textbf{16.2} \\
%         %\midrule
%         %\midrule
%         %Clean Model & Clean-FCOS* & 34.9 & 34.9 & 34.9 & 34.9 & 34.9 \\
%         %Clean Model & Clean-Faster R-CNN* & 45.0 & 45.0 & 45.0 & 45.0 & 45.0 \\
%         \bottomrule
%         \end{tabular}
% 	\label{tab:combinednoise}
% \end{table*}

\begin{table}
	\centering
	\caption{Performance of baseline methods and our DN-TOD on the proposed AI-TOD-v2.0 {\tt val set} with different levels of mixed noise. * means using RFLA~\citep{rfla} in the label assignment.}
    \resizebox{\linewidth}{!}{
	\begin{tabular}{l | c c c c c}%四个c代表该表一共四列，内容全部居中
        \toprule
        \multirow{2}{*}{Method}& \multicolumn{5}{c}{Noise Level} \\%表格宽度参数采用*代表自动宽度
         & 0\% & 10\% & 20\% & 30\% & 40\% \\
        \midrule%第二道横线 
        FCOS* & 34.9 & 33.0 & 28.5 & 24.9 & 12.7 \\
        %\cline{2-6}
        DN-TOD & \textbf{38.1} & \textbf{34.2} & \textbf{31.4} & \textbf{27.2} & \textbf{17.6} \\
        \midrule
        Faster R-CNN* & 45.0 & 43.7 & 33.0 & 24.2 & 12.8 \\
        %\cline{2-6}
        DN-TOD & \textbf{45.5} & \textbf{45.3} & \textbf{38.1} & \textbf{29.1} & \textbf{16.2} \\
        %\midrule
        %\midrule
        %Clean Model & Clean-FCOS* & 34.9 & 34.9 & 34.9 & 34.9 & 34.9 \\
        %Clean Model & Clean-Faster R-CNN* & 45.0 & 45.0 & 45.0 & 45.0 & 45.0 \\
        \bottomrule
        \end{tabular}}
	\label{tab:combinednoise}
\end{table}

\begin{table}
	\centering
	\caption{Generalization to real-world tiny object detection datasets. We train on the AI-TOD {\tt train set} and evaluate on the AI-TOD-v2.0 {\tt val set}. * means using RFLA~\citep{rfla} in the label assignment.}
    \resizebox{\linewidth}{!}{
	\begin{tabular}{l | c c c c c c}
        \toprule
        Method & mAP & $\rm{AP_{0.5}}$ & $\rm{AP_{vt}}$ & $\rm{AP_{t}}$ & $\rm{AP_{s}}$ & $\rm{AP_{m}}$ \\
        \midrule
        Noisy-FCOS* & 14.8 & 32.6 & 6.7 & 16.9 & 16.3 & 19.1 \\
        % \cline{2-7}
        DN-TOD & \textbf{15.5} & \textbf{34.9} & \textbf{6.8} & \textbf{17.8} & \textbf{16.3} & \textbf{19.4} \\
        % \midrule
        \midrule
        Clean-FCOS* & 15.8 & 34.9 & 7.1 & 18.1 & 16.9 & 19.7 \\
        \bottomrule
        \end{tabular}}
	\label{tab:realnoise}
\end{table}

\textbf{Mixed Noise.} In practical applications, label noise typically manifests in a mixed manner. To better assess the performance of our method in more practical scenarios, we conduct further experiments combining class shifts and box noises. The experimental results, as shown in Table~\ref{tab:combinednoise}, indicate that our method exhibits substantial improvements compared to vanilla FCOS* and Faster R-CNN* models. Delving into details, the patterns observed in detection results amid mixed noise display certain similarities to those observed under individual noise, while also manifesting distinctions. Similarly, one-stage methods commonly yield results lower than two-stage methods under low levels of label noise, however, as the noise level increases, the two-stage method degrades faster. Differently, the improvement brought by DN-TOD under mixed noise is not as prominent as the improvement brought by each module under individual noise. When a mixture of noise happens, challenges posed to the model are not limited to a combination of issues under different noises.

\textbf{Real-world Noise.} In the real world, the label noise in object detection datasets is complex and hard to predict. In the field of tiny object detection in aerial images, we currently have a natural pair of \textit{noisy vs. less noisy} dataset, namely AI-TOD \textit{vs.} AI-TOD-v2.0. AI-TOD-v2.0 is re-annotated based on the AI-TOD, where they share the same images while AI-TOD-v2.0 has annotations of higher quality~\citep{aitodv2_2022_isprs}.
To verify the method's transferability to tackle real-world noise, we train the DN-TOD on AI-TOD and validate it on AI-TOD-v2.0. Results are shown in Table~\ref{tab:realnoise}, the performance of DN-TOD trained on AI-TOD almost aligns with the FCOS* trained on AI-TOD-v2.0, which implies the strong robustness of DN-TOD against real-world label noise in tiny object detection. 

\subsection{Ablation Study}
\begin{table*}[t]\vspace{-3mm}
% subfloat a - BackBone Architecture
\centering
\caption{Ablations. We train on noisy AI-TOD-v2.0 \texttt{train set} under 20\% noise rate, test on the clean \texttt{val set}.}
\subfloat[Individual effectiveness of methods in TLS. (Box noise) \label{tab:dss}]{
\tablestyle{4pt}{1.05}\begin{tabular}{ccccccccc}  
	\toprule
	TLR & RBR  & mAP & $\rm{AP_{0.5}}$ & $\rm{AP_{0.75}}$ & $\rm{AP_{vt}}$ & $\rm{AP_{t}}$ & $\rm{AP_{s}}$ & $\rm{AP_{m}}$ \\
	\midrule
	 & & 11.4 & 30.2 & 5.0 & 5.0 & 13.5 & 11.5 & 14.4 \\
	 \checkmark & & 12.8 & 33.7 & 6.3 & 4.9 & 14.9 & 13.1 & 16.0 \\
	 & \checkmark & 12.6 & 33.2 & 6.0 & 5.1 & 14.6 & 12.9 & 16.5 \\
	 \checkmark & \checkmark & \textbf{14.0} & \textbf{37.4} & \textbf{6.4} & \textbf{5.6}  & \textbf{15.6} & \textbf{15.8} & \textbf{16.7}\\
     \midrule
    \multicolumn{2}{c}{Noisy-FCOS*} & 11.4 & 30.2 & 5.0 & 5.0 & 13.5 & 11.5 & 14.4\\
    \multicolumn{2}{c}{Clean-FCOS*} & 15.8 & 34.9 & 11.2 & 7.1 & 18.1 & 16.9 & 19.7\\
	\bottomrule
	\end{tabular}}\hspace{3mm}
% subfloat b - Multinomial vs Independent Masks
\subfloat[Effects of parameter $T$ in updating DCM. (Class noise) \label{tab:T}]{
\tablestyle{4pt}{1.05}\begin{tabular}{cccccccc}  
	\toprule
	$ T $  & mAP & $\rm{AP_{0.5}}$ & $\rm{AP_{0.75}}$ & $\rm{AP_{vt}}$ & $\rm{AP_{t}}$ & $\rm{AP_{s}}$ & $\rm{AP_{m}}$ \\
	\midrule
	  50 & 14.7 & 31.6 & 10.8 & 6.6 & 17.1 & 15.4 & 18.6 \\
	100 & \textbf{15.6} & \textbf{35.1} & \textbf{10.9} & 6.9 & \textbf{17.7} & \textbf{16.6} & \textbf{19.4} \\
        500 & 15.4 & 35.1 & 10.5 & 6.8 & 17.6 & 16.0 & 18.1 \\
        1000 & 15.3 & 33.5 & 10.9 & \textbf{7.1} & 17.5 & 16.4 & 18.5 \\
	  \midrule
        Noisy-FCOS* & 14.9 & 32.6 & 10.9 & 6.5 & 17.0 & 16.3 & 18.2 \\
        Clean-FCOS* & 15.8 & 34.9 & 11.2 & 7.1 & 18.1 & 16.9 & 19.7 \\
	\bottomrule
	\end{tabular}}\\
% subfloat c - RoIAlign (ResNet-50-C4)
\subfloat[Effects of different $\alpha$ for balancing factors in TLR. (Box noise)\label{tab:alpha}]{
\tablestyle{4pt}{1.25}\begin{tabular}{cccccccc}  
	\toprule
	$\alpha$  & mAP & $\rm{AP_{0.5}}$ & $\rm{AP_{0.75}}$ & $\rm{AP_{vt}}$ & $\rm{AP_{t}}$ & $\rm{AP_{s}}$ & $\rm{AP_{m}}$ \\
	\midrule
	  0.2 & 13.9 & 37.2 & \textbf{6.5} & 5.6 & 15.5 & 15.5 & \textbf{17.5} \\
	0.5 & \textbf{14.0} & \textbf{37.4} & 6.4 & 5.6 & \textbf{15.6} & \textbf{15.8} & 16.7 \\
	  0.8 & 13.5 & 36.6 & 6.1 & \textbf{6.1} & 15.4 & 14.6 & 17.3 \\
	  \midrule
        Noisy-FCOS* & 11.4 & 30.2 & 5.0 & 5.0 & 13.5 & 11.5 & 14.4\\
        Clean-FCOS* & 15.8 & 34.9 & 11.2 & 7.1 & 18.1 & 16.9 & 19.7\\
	\bottomrule
	\end{tabular}}\hspace{3mm}
% subfloat d - Effects of parameters 
\subfloat[Effects of $w_1$ for deciding \textit{gt}'s weight in RBR. (Box noise)\label{tab:omega}]{
\tablestyle{4pt}{1.05}\begin{tabular}{cccccccc}  
	\toprule
	$w_1$  & mAP & $\rm{AP_{0.5}}$ & $\rm{AP_{0.75}}$ & $\rm{AP_{vt}}$ & $\rm{AP_{t}}$ & $\rm{AP_{s}}$ & $\rm{AP_{m}}$ \\
	\midrule
	  0.5 & 13.7 & 36.7 & 6.4 & 5.2 & 15.5 & 15.7 & 16.2 \\
	1.0 & \textbf{14.0} & \textbf{37.4} & \textbf{6.4} & 5.6 & \textbf{15.6} & \textbf{15.8} & 16.7 \\
	  2.0 & 13.7 & 36.1 & 6.1 & 5.1 & 15.6 & 14.9 & 16.0 \\
        5.0 & 13.4 & 35.8 & 5.9 & \textbf{5.7} & 15.3 & 14.2 & \textbf{17.1} \\
	  \midrule
        Noisy-FCOS* & 11.4 & 30.2 & 5.0 & 5.0 & 13.5 & 11.5 & 14.4\\
        Clean-FCOS* & 15.8 & 34.9 & 11.2 & 7.1 & 18.1 & 16.9 & 19.7\\
	\bottomrule
	\end{tabular}}
% main caption
\label{tab:ablations}\vspace{-3mm}
\end{table*}

This section investigates the contribution of key designs and the selection of hyper-parameters. First, we verify the effects of the Trend-guided Label Reweighting and Recurrent Bounding box Regeneration, respectively. Then, we study the influence of the selection of different hyper-parameters. Note that ablations are performed on the AI-TOD-v2.0 dataset.

\textbf{Trend-guided Label Reweighting (TLR).} As shown in Table~\ref{tab:dss}, it is evident that merely utilizing the TLR module can already boost performance. For instance, we achieve 3\% $\rm{AP_{0.5}}$ improvements under 20\% box noise level. We believe that the TLR module's ability to rectify errors induced by the label assignment can improve the robustness against box annotation offsets. 

\begin{figure*}[t]
    \centering
    \includegraphics[width=0.99\linewidth]{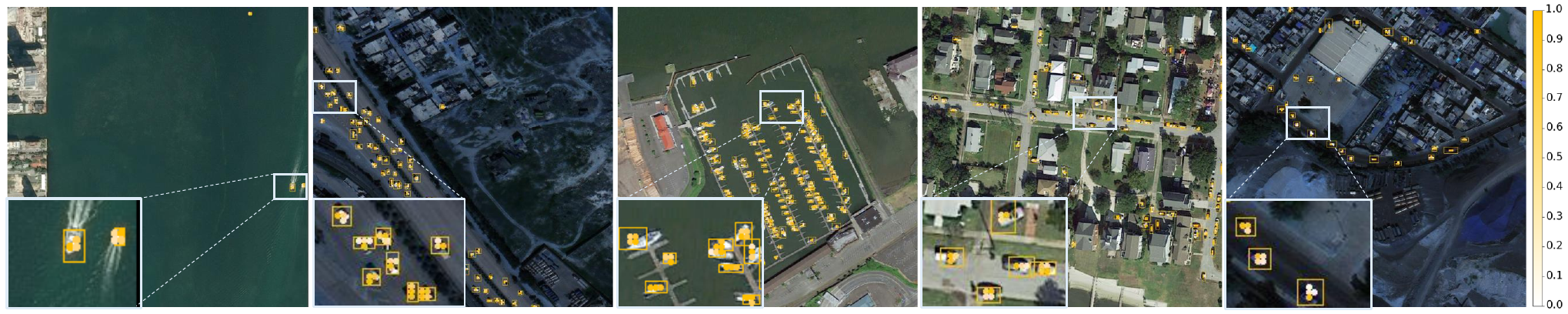}
    \caption{Visualization of the sample weights by TLR. The points depict the cleanliness weighted coefficients generated by the TLR for each positive sample. Here, we demonstrate the cleanliness weighted coefficients under the inaccurate bounding boxes (orange box). It can be observed that the TLR strengthens the positive sample points at relatively clean positions.}
    \label{fig:tlr}
    %\vspace{\fixedvskip}
\end{figure*}

\begin{figure*}[t]
    \centering
    \includegraphics[width=0.97\linewidth]{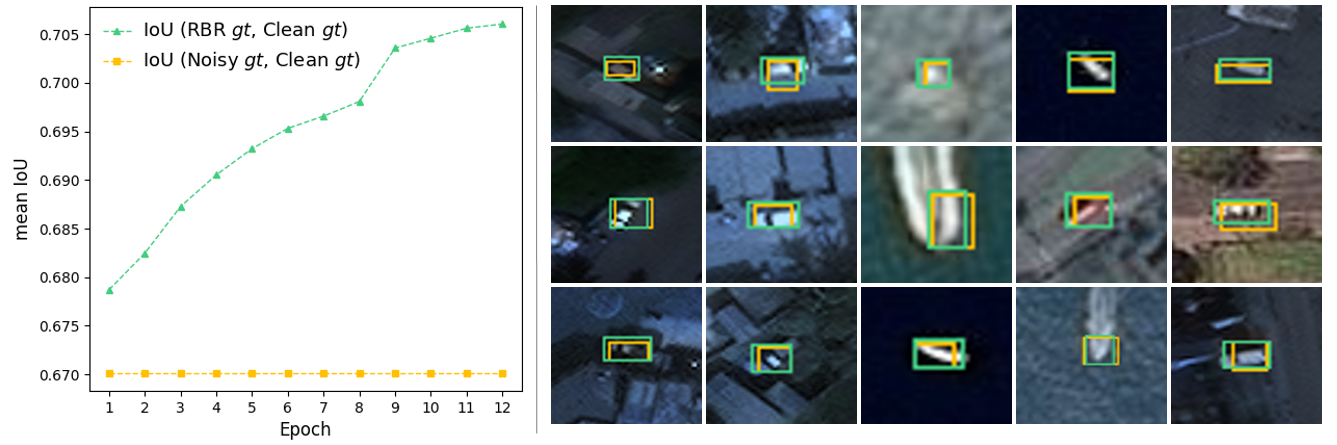}
    \caption{The green box represents bounding boxes generated by the RBR module, and the orange box denotes noisy bounding boxes. The curve on the left indicates that as training progresses, the bounding boxes generated by the RBR module will gradually approach the clean bounding boxes.}
    \label{fig:rbrgeneration}
    %\vspace{\fixedvskip}
\end{figure*}

\textbf{Recurrent Bounding box Regeneration (RBR).} The RBR module can further improve the network's robustness against bounding box noise. In particular, under 20\% noise level, the RBR module can boost the $\rm{AP_{0.5}}$ performance from 33.7 to 37.4 (Table~\ref{tab:dss}). This further confirms the RBR's ability to provide a cleaner regression target via the weighted bounding box ensembling.

\textbf{$T$ in CLC.} $T$ represents the period in the CLC module, which is similar to the notion of the ``sliding window length'' in the time series operation. A large $T$ will take samples across a longer period previous to the current iteration, while a small $T$ will consider less previous samples when calculating the transition matrix. As shown in Table~\ref{tab:T}, when $T$ is set to 100, the best performance is obtained. Checking a series of values from 50 to 1000, we find that the robustness can be well-maintained when we set $T$ larger. Generally, as $T$ increases, the samples are more representative of the whole dataset, while the current training status cannot the well-reflected. Conversely, when $T$ is smaller, the instability resulting from fluctuations in training has a more significant impact. Therefore, to strike a balance between stability and the ability to express current status, we suggest setting $T$ to 100.

\textbf{$\alpha$ in TLR.} We design a balancing factor $\alpha$ to alleviate the unstable re-weighting of the TLR in the early stages of training. In Table~\ref{tab:alpha}, we keep all other parameters fixed and experimentally demonstrate that the difference of $\alpha$ has a relatively minor impact on the results. However, to ensure stability in the early stages, it is recommended to set the value of $\alpha$ at 0.5 to achieve relatively favorable results.

\textbf{$w_1$ in RBR.} To avoid drastic fluctuations in the supervisory signal during each round, we introduced the $w_1$ to incorporate the \textit{gt} as part of the refactored box. As demonstrated in Table~\ref{tab:omega}, experimental results indicate that under 20\% noise level, setting $w_1$ to 1.0 yields relatively favorable outcomes. Given that $w_1$ represents the weighting factor of the \textit{gt} signal when the dataset exhibits higher levels of noise, we lean towards selecting smaller values for $w_1$. It will provide the network with a cleaner supervisory signal.

\begin{table}
	\centering
	\caption{Performance of baseline methods and our proposed method on the DOTA-v2.0 {\tt val set}. * means using RFLA~\citep{rfla} in the label assignment.}
	\begin{tabular}{l | c c c c c}%四个c代表该表一共四列，内容全部居中
        \toprule
        \multirow{2}{*}{Method}& \multicolumn{5}{c}{Noise Level} \\%表格宽度参数采用*代表自动宽度
         & 0\% & 10\% & 20\% & 30\% & 40\% \\
        \midrule%第二道横线 
        FCOS* & 27.8 & 24.5 & 19.1 & 10.7 & 4.4 \\
        % \cline{2-7}
          DN-TOD & \textbf{28.1} & \textbf{25.6} & \textbf{20.0} & \textbf{10.9} & \textbf{5.0} \\
        % \midrule
        % \midrule
        % Clean-FCOS* & 27.8 & 27.8 & 27.8 & 27.8 & 27.8 \\
        \bottomrule
        \end{tabular}
	\label{tab:dotacombine}
\end{table}

\subsection{Generalization to Aerial Object Detection}
Our DN-TOD can also be adapted to tackle the label noise in generic aerial object detection datasets which contain a considerable number of tiny objects. For instance, we select the latest version of DOTA (\textit{i.e.}, DOTA-v2.0), which is featured by its massive number of tiny objects for verification. Similarly, we choose FCOS with RFLA as the baseline detector and build the proposed DN-TOD on it. Comparisons with baseline detectors are reported
in Table~\ref{tab:dotacombine}. Detection results show the improvements in the DOTA-v2.0 dataset under the mixed noise situation.

\subsection{Visual Analysis}
We conduct a group of analytical experiments to demonstrate our methods can identify noisy samples and provide cleaner box supervision for training. First, we visualize the positive sample weights generated by the TLR module (Fig.~\ref{fig:tlr}). It can be noticed that even if bounding boxes deviate from real objects, the TLR module can pose higher weights to cleaner samples falling on the objects' main body. As a consequence, TLR can help alleviate classification bias caused by inaccurately annotated bounding boxes during label assignment.
Second, we visualize the bounding boxes generated by the RBR module in each round and their IoU (Intersection over Union) with the clean \textit{gt} (Fig.\ref{fig:rbrgeneration}). We can observe that, compared to noisy annotations, new boxes generated by RBR are closer to tiny objects' main body, which greatly rectifies the regression target.

% \input{dotav2}

% discussion
\begin{figure*}[t]
    \centering
    \includegraphics[width=0.98\linewidth]{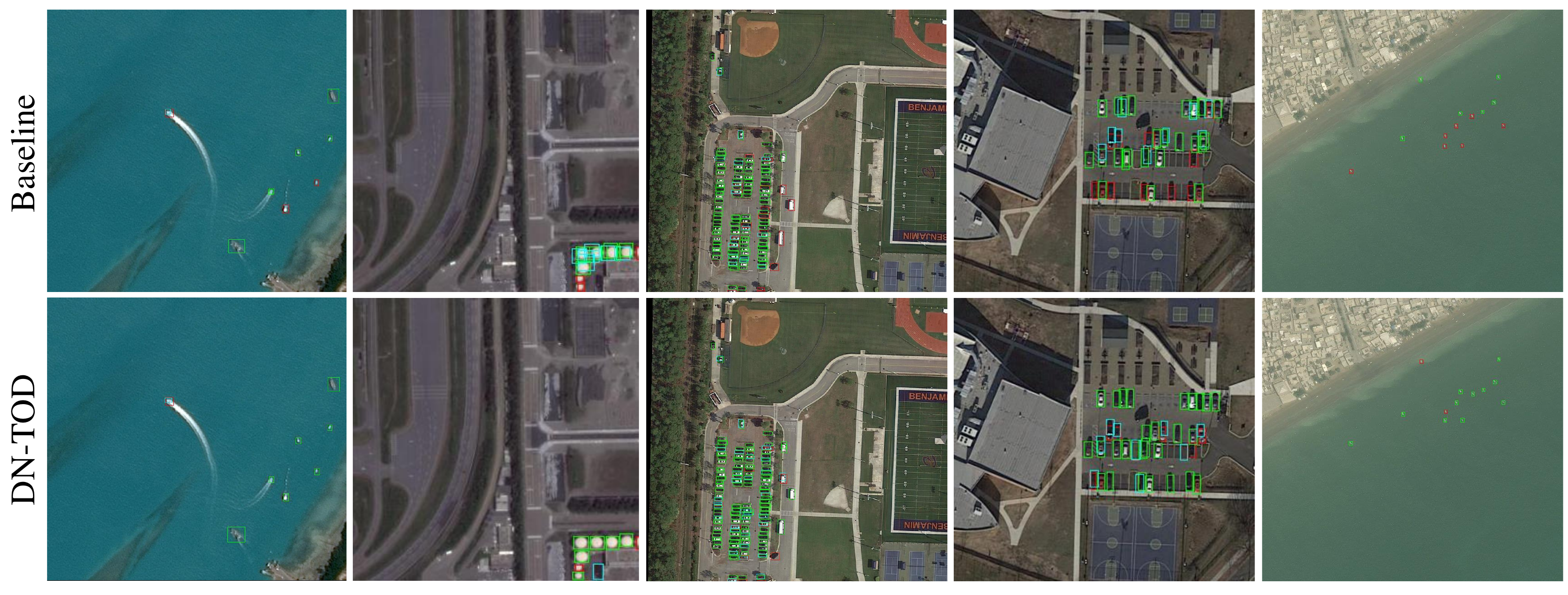}
    \caption{Visualization results on AI-TOD-v2.0 under 30\% noise. The first row is the result of FCOS* and the second row is the result of DN-TOD. Green boxes denote true positive predictions, red boxes denote false negative predictions, and blue boxes denote false positive predictions.}
    \label{fig:cpmpare}
    %\vspace{\fixedvskip}
\end{figure*}

\section{Discussion}
\label{discussion}
Tiny objects are ubiquitous in aerial imagery, and the precise recognition and localization of these objects are crucial to numerous remote sensing applications~\citep{DIOR_2019_ISPRS,NWPU_VHR-10_2014_ISPRSJ}, like aerial-based traffic surveillance, wildlife observation~\citep{mammals_2018_rse,benchmar_wildlife_2016_isprs}, and marine debris detection~\citep{debris_2023_iscience}. Due to the lack of appearance information and the large quantity per image, tiny objects tend to be nosily labeled, thus worsening the already unsatisfactory detection performance. 

% how we respond to the label noise in tod
This motivates us to specifically study the robust tiny object detection in aerial imagery under noisy label supervision. We, for the first time, perform a systematic study of the effects of different types of noise on tiny object detection. Identifying the model's vulnerability to class shifts and inaccurate bounding box annotations, we further propose a DN-TOD that pursues the expected robustness. However, we face significant challenges during this process.

% how we tackle these challenges, our solutions and advantages
For class shifts, we observe that the intrinsic class imbalanced distribution of tiny objects can severely affect label correction. Previous works like co-teaching~\citep{coteaching} and hard-threshold filtering are prone to retain more samples for frequent classes, deteriorating the learning of severely class-imbalanced tiny objects~\citep{labelnoiseclassimbalance}. In response, we design a CLC scheme, which separates the status updating and noisy sample filtering of different classes. Experiments on the noisy AI-TOD-v2.0 verify the CLC's resistance to different levels of class shifts. 
For box noise, existing state-of-the-art methods (\textit{e.g.}, OA-MIL~\citep{OAMIL} and SSD-Det~\citep{SSDdet}) yield sub-optimal performance on tiny object datasets. This can be largely attributed to the bias during the process of proposal bag construction. In this process, clean samples are identified based on classification confidence, however, in aerial images, large objects tend to yield higher classification confidence, suppressing tiny objects under this paradigm. By contrast, we use the learning trend of positive samples as a substitute for the scale-biased classification confidence, to reweight samples and regenerate cleaner box targets. Extensive quantitative and qualitative experiments demonstrate the superiority of our TLS in handling tiny object box noise.

% limitations
Despite these advances, there remain some pending challenges for this subject. First, since the DN-TOD is designed for tiny object detection, the improvement on multi-scale objects is not as obvious as that on tiny objects, which can be inferred from the results on DOTA-v2.0 (Table~\ref{tab:dotacombine}). We propose to specifically study the label noise in tiny objects since it is more severe and common. However, both the robust detection of tiny objects and larger objects play important roles in aerial image interpretation. 
Second, the improvement from our method is weakened when confronted with mixed noise compared to the performance under single noise. We attribute this discrepancy to the heightened impact of mixed noise on the network. It cannot be simply regarded as an additive effect of different noises; hence, the straightforward combination of different strategies does not result in a significant enhancement.

% future works, three levels: method, pipeline, task, 
Moreover, our experimental results and analyses suggest that future works may benefit from the following aspects. First, an improved method can be developed to better tackle tiny object detection under any mixed type of noise. Second, a unified pipeline can be considered that simultaneously addresses all types of noise for different-sized objects in aerial imagery, simplifying the tuning cost when deploying. 
Third, although the label noise issue is pronounced for the tiny object detection task, it can be commonly found in other remote sensing interpretation tasks, like scene classification, oriented object detection, and semantic segmentation in aerial images. Thus, the ideas in this paper can be further expanded to benefit research in these fields.

% conclusion
\section{Conclusion}
\label{conclusion}
%revised by xc
Tiny object detection becomes more challenging when coupled with inaccurate annotations, yet precisely detecting tiny objects under such noisy situations is more in line with real-world scenarios. In this paper, we have investigated robust tiny object detection under various types of noisy label supervision by first discovering the network's sensitivity to class shifts and inaccurate box annotations and then proposing a DN-TOD to mitigate their effects. The DN-TOD is characterized by a CLC module for class shifts and a TLS module for inaccurate bounding boxes. CLC effectively identifies and filters class-shifted noisy samples by a class-aware DCM. TLS alleviates the classification and regression bias induced by box noise via TLR and RBR, respectively. Experiments on the synthetic and real-world noisy tiny object detection datasets demonstrate the superiority and robustness of our method when facing label noise.

\section*{Acknowledgements}
The research was partially supported by the National Natural Science Foundation of China (NSFC) under Grants 62271355. The numerical calculations were conducted on the supercomputing system in the Supercomputing Center, Wuhan University. 

\printcredits

\bibliography{Jinwang-Papers, chang}
%\printbibliography
\end{sloppypar}
\end{document}